%% file: main.tex
\documentclass{article}
\usepackage{spconf,amsmath,amssymb,amsthm,graphicx,booktabs,multirow,xcolor,enumitem}
\usepackage{algorithm,algpseudocode}
\usepackage{tcolorbox}
\usepackage{cuted}
\usepackage{eso-pic}
\definecolor{lightpurple}{rgb}{0.862,0.918,0.992}
\definecolor{githubpink}{RGB}{255,0,128}
\newtcolorbox{thmbox}{colback=lightpurple,colframe=lightpurple,sharp corners=all,boxrule=0mm,boxsep=0.5mm,left=1.5mm,right=1.5mm,top=1.2mm,bottom=1.2mm}
\usepackage[hidelinks]{hyperref}
\usepackage[protrusion=true,expansion=true]{microtype}

\newcommand{\rulefix}{\dimexpr-.5\aboverulesep-.5\belowrulesep-.5\cmidrulewidth\relax}

\makeatletter
\providecommand{\@setmarks}{}

\makeatother
\newcommand{\Score}{\operatorname{Score}}

\title{Count Evidence, Not Sentences: Tempered Evidence Fusion of LLM Judgments for Long-Text Value Measurement}

\name{\fontsize{10pt}{12pt}\selectfont\begin{tabular}{c}Yuhe Wu$^{1,*}$, Rui Qian$^{2,*}$, Guangyu Wang$^{1,*}$, Yuran Chen$^{3}$, Yuanchao Zhu$^{4}$, Junjie Yang$^{5}$\\ Zhengheng Li$^{6}$, Jiulin Cai$^{7}$, Tianyi Zhang$^{8}$, Zihan Dong$^{9}$, Jiaxin Liu$^{1}$, Yujie Chen$^{10}$, Guang Zhang$^{1,\dagger}$\end{tabular}\thanks{$^{*}$Equal contribution $^{\dagger}$Corresponding author}}
\address{\fontsize{10pt}{12pt}\selectfont $^{1}$HKUST(GZ)\quad $^{2}$FDU\quad 
$^{3}$DUFE\quad $^{4}$UESTC\quad $^{5}$UMD\\ \fontsize{11pt}{12pt}\selectfont $^{6}$SEU\quad $^{7}$USTC\quad $^{8}$Independent\quad $^{9}$Georgia Tech\quad $^{10}$CUHK(SZ)}

\newcommand{\titleemail}{%
  \AddToShipoutPictureFG*{%
    \AtPageUpperLeft{%
      \raisebox{-212pt}[0pt][0pt]{%
        \makebox[\paperwidth][c]{\fontsize{10pt}{12pt}\selectfont\texttt{\{yuheewuu,qiianruii\}@gmail.com, \small\texttt{{guangzhang}@hkust-gz.edu.cn}}}%
      }%
    }%
  }%
}

\begin{document}
\ninept
\titleemail
\maketitle
  
\begin{abstract}
Large language models (LLMs) are increasingly used to measure public value orientations from 
long social media posts, yet such posts often mix background, quotations, concessions, and 
only a few stance-bearing sentences. Existing approaches either ask the model to predict a 
document-level label directly, which can be overconfident, or aggregate sentence-level 
predictions by majority or soft voting, which treat uncertain and decisive sentences as 
equally informative. We formulate long-text value measurement as a decision-fusion problem 
and propose Tempered Evidence Fusion (TEF), a training-free rule that weights each sentence's 
log-odds by its normalized information gain, as derived from a generalized Bayesian posterior. 
This makes the fused score nearly vanish for uncertain sentences while preserving the 
Bayes-optimal weight of decisive evidence. We further introduce Multi-event Insight 
Network Dimensions (MIND), a benchmark of 8,358 Chinese and English posts spanning 
five years of public events and six value dimensions. On MIND, TEF outperforms the strongest 
baseline among Direct, Majority Vote, and Soft Vote by an average of 4.5 accuracy points and 
4.6 macro-F1 points across five LLMs and two languages. MIND dataset and code are available at 
\href{https://github.com/Kzczc/ICASSP2027-TEF}{\textcolor{githubpink}{GitHub}}.
\end{abstract}

\begin{keywords}
Decision fusion, large language models, uncertainty, calibration, computational social science
\end{keywords}

\input{sections/1_intro}
\input{sections/2_related}
\input{sections/3_method}
\input{sections/4_benchmark}
\input{sections/5_experiments}
\input{sections/6_conclusion}
\input{sections/7_ethics}

\bibliographystyle{IEEEbib}
\clearpage
\bibliography{refs}
\end{document}

%% file: sections/1_intro.tex
\section{Introduction}
\label{sec:intro}

Public value orientations are key latent variables in computational social science, capturing how people position 
themselves on issues such as economic regulation, climate policy, or cultural change. Social media makes these 
orientations observable through text at scale~\cite{conover2011political,lee2025semantic}. Yet this scale, together 
with the subjectivity of human annotation, has motivated growing interest in using LLMs as measurement 
instruments~\cite{pangakis2025keeping,islam2025uncovering}. Existing LLM-based pipelines usually either ask the model 
to read the whole post and return a document-level label~\cite{gilardi2023chatgpt,ziems2024can,tornberg2024large}, or split the post into sentences and aggregate 
sentence-level predictions by voting~\cite{tsirmpas2024longtexts}.

\begin{figure}[t]
\centering
\includegraphics[width=\columnwidth]{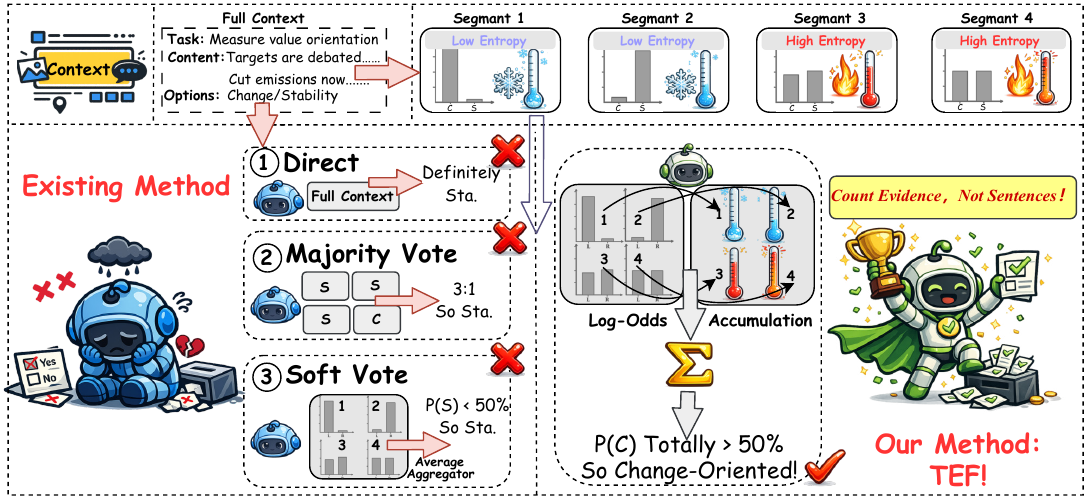}
\vspace{-0.6cm}
\caption{Long-text value measurement requires counting informative evidence rather than treating all sentences equally. \textbf{LEFT:} Direct prediction collapses mixed content into one label, while voting can let many uncertain sentences outweigh a few decisive ones. \textbf{RIGHT:} TEF weights sentence log-odds by normalized information gain, discounting uncertain judgments so decisive evidence drives the document-level orientation.}
\label{fig:teaser}
\vspace{-0.6cm}
\end{figure}

However, we observe that little research has examined how sentence-level LLM judgments should be fused when evidence 
is unevenly distributed across a long post. In Fig.~\ref{fig:teaser}, long social media posts often mix background 
information, quotations, and concessions, with only a few stance-bearing sentences. In such cases, direct prediction 
can collapse heterogeneous evidence into one overconfident label, while majority or soft voting can allow  uncertain 
sentences to outvote a few decisive ones. This leads us to ask: \emph{How can sentence-level judgments be combined so 
that decisive evidence dominates the final decision while uncertain sentences are discounted?}

Bearing this in mind, we formulate long-text value measurement as a decision-fusion 
problem, where each sentence acts as a local detector that emits a soft decision. 
Under conditional independence and calibrated sentence posteriors, the Bayes-optimal 
document-level rule sums sentence log-likelihood ratios. This observation places 
existing voting rules on a common scale: majority voting keeps only the sign of each 
sentence judgment, and soft voting uses a bounded surrogate that underweights highly 
decisive sentences. At the same time, directly summing LLM logits is unreliable 
because next-token probabilities are only approximately calibrated, especially near 
the point of maximal uncertainty.

To this end, we present \textbf{T}empered \textbf{E}vidence \textbf{F}usion 
(\textbf{\textit{TEF}}), a training-free fusion rule that tempers each sentence's 
log-odds by its normalized information gain, as derived from a generalized Bayesian 
posterior~\cite{bissiri2016general,grunwald2017inconsistency}. The resulting score 
nearly vanishes for uncertain sentences but preserves the Bayes-optimal weight of 
decisive evidence, yielding a soft censoring effect similar in spirit to censoring 
rules in distributed detection~\cite{rago1996censoring}. In Fig.~\ref{fig:teaser} 
(right), TEF counts evidence rather than sentences, allowing a few informative 
sentences to determine the document-level orientation.

Our contributions are threefold. (i) We cast long-text value measurement with LLMs 
as a decision-fusion problem and derive TEF from a tempered Bayesian posterior, 
showing that it suppresses uncertain sentences while preserving the weight of 
decisive evidence. (ii) We introduce MIND, a five-year, cross-lingual benchmark 
of 8,358 posts over six value dimensions. (iii) We show that TEF outperforms the 
strongest of the Direct, Majority Vote, and Soft Vote baselines by 4.5 accuracy 
points and 4.6 macro-F1 points on average across five LLMs and two languages, 
yielding better-calibrated estimates.



%% file: sections/2_related.tex
\section{Related Work}
\label{sec:related}

LLM-based stance and ideology measurement has expanded from short-form texts such as tweets to longer, 
multi-issue documents~\cite{saha2024stance,akash2025open,bhattacharya2025rethinking}. In parallel, 
the reliability of LLM confidence has been studied from multiple perspectives, including token-level 
probabilities~\cite{jiang2021calibration,kadavath2022know}, verbalized confidence~\cite{tian2023just,shorinwa2025survey}, and 
aggregation methods beyond majority voting~\cite{ai2025beyond}. In signal processing, fusing local 
decisions is a classical problem~\cite{tenney1981detection}: log-linear opinion pools combine probabilistic 
judgments~\cite{genest1986combining}, the Chair--Varshney rule weights local decisions by their 
reliabilities~\cite{chair1986optimal}, and censoring schemes transmit only observations with 
informative likelihood ratios~\cite{rago1996censoring}. Despite these connections, long-text 
measurement with LLMs still largely relies on equal-weight voting. In contrast, we formulate 
long-text value measurement as sentence-level decision fusion and derive an evidence-aware 
fusion rule, TEF, that discounts uncertain sentences while preserving the weight of decisive ones.

%% file: sections/3_method.tex
\section{Proposed TEF}
\label{sec:methodology}

\noindent\textbf{Problem Formulation.} We formulate long-text value measurement as a decision-fusion problem.
The goal is to infer a latent social variable from a collection of textual segments whose evidential strength may vary substantially within the same document.
Let $\mathcal{X}$ denote the input space and let $\mathcal{Y}=\{c_1,\ldots,c_K\}$ be the set of target measurement categories.
For an observation $X\in\mathcal{X}$, the latent measurement variable $Y$ takes values in $\mathcal{Y}$.
We segment each long text into $N$ sentence-level units,
\begin{equation}
\vspace{-0.1cm}
X \mapsto \mathcal{D}(X)=\{D_1,D_2,\ldots,D_N\},
\end{equation}
where $D_i$ denotes the $i$-th segment.
This granularity preserves local semantic coherence while keeping inference computationally manageable. 
For each segment $D_i$, a language model $\mathcal{M}$ returns a probability distribution over the measurement categories, denoted by
$\mathbf{p}_i=\mathcal{M}(D_i)=\big(p_i^{(1)},\ldots,p_i^{(K)}\big)$,
where each entry $p_i^{(k)}=P_{\mathcal{M}}(Y=c_k\mid D_i)$
and $\mathbf{p}_i\in\Delta^{K-1}$.
The distribution encodes both a local prediction $\hat{y}_i=\arg\max_k p_i^{(k)}$ and its uncertainty, measured by entropy
$H(\mathbf{p}_i)=-\sum_{k=1}^{K}p_i^{(k)}\log p_i^{(k)}.$
Common aggregation rules discard part of this probabilistic evidence.
Majority voting uses only the segment-level hard labels, while soft voting averages probability vectors with equal weight,
\begin{equation}
\vspace{-0.2cm}
\hat{Y}_{\mathrm{MV}}=\mathrm{mode}\big(\{\hat{y}_i\}_{i=1}^{N}\big),
\qquad
\hat{Y}_{\mathrm{SV}}=\arg\max_{c\in\mathcal{Y}}\frac{1}{N}\sum_{i=1}^{N}p_i^{(c)}.
\end{equation}
Both rules treat all segments as equally reliable.
This is problematic for scientific measurement from long texts, where decisive evidence may appear in only a few segments and many other segments may be ambiguous, rhetorical, or weakly relevant. TEF instead aggregates reliability-weighted evidence.
Let $w_i=\omega(\mathbf{p}_i)\in[0,1]$ be the reliability assigned to segment $D_i$, and let $g:[0,1]\to\mathbb{R}$ be a scoring transformation.
The document-level measurement is $\hat{Y}=\arg\max_{c\in\mathcal{Y}}\sum_{i=1}^{N}w_i\,g\big(p_i^{(c)}\big).$
The following paragraphs specify $\omega$ and $g$, which together define the fusion rule of TEF.
\vspace{+0.2cm}

\noindent\textbf{Reliability Weighting.}
TEF instantiates the preceding framework with entropy-based reliability weighting and one-vs-rest log-odds evidence.
Algorithm~\ref{alg:framework} gives the complete procedure.
We use normalized information gain as the segment reliability,$
w_i=1-\frac{H(\mathbf{p}_i)}{\log K}\in[0,1].$
This weight is close to one for concentrated distributions and close to zero for near-uniform distributions.
Thus confident segments contribute strongly, while uncertain segments are softly censored.

\noindent\textbf{Evidence Aggregation.}
For each category $c$, TEF transforms the segment posterior into a one-vs-rest log-odds score,
$g\big(p_i^{(c)}\big)=\log\frac{p_i^{(c)}}{1-p_i^{(c)}}.$
The aggregated evidence for category $c$ is
\vspace{-0.3cm}
\begin{equation}
\Score(c)=\sum_{i=1}^{N}w_i\log\frac{p_i^{(c)}}{1-p_i^{(c)}}.
\label{eq:score}
\vspace{-0.3cm}
\end{equation}
The final measurement outcome is $\hat{Y}=\arg\max_{c\in\mathcal{Y}}\Score(c)$.
In implementation, probabilities are clipped to $[\epsilon,1-\epsilon]$ before applying the logit, and log-odds values are clipped to $[-M,M]$.
Unless otherwise stated, we set $\epsilon=10^{-6}$ and $M=10$. Unlike majority voting, TEF does not reduce $\mathbf{p}_i$ to a hard label before aggregation.
Unlike soft voting, TEF does not assume that all segments are equally informative.
It therefore preserves distributional information while reducing the impact of noisy or ambiguous segments.

\begin{algorithm}[htbp]
\caption{Tempered Evidence Fusion}
\footnotesize
\label{alg:framework}
\begin{algorithmic}[1]
\Require Text $X$, model $\mathcal{M}$, categories $\mathcal{Y}=\{c_1,\ldots,c_K\}$, clip bound $M$, stability constant $\epsilon$.
\Ensure Measurement outcome $\hat{Y}$
\State $\{D_1,\ldots,D_N\}\gets\mathcal{D}(X)$ \Comment{Segment into sentences}
\State Initialize $\Score(c)\gets0$ for all $c\in\mathcal{Y}$
\For{$i=1$ \textbf{to} $N$}
    \State $\mathbf{p}_i\gets\mathcal{M}(D_i)$ \Comment{Segment-level distribution}
    \State $H_i\gets-\sum_{k=1}^{K}p_i^{(k)}\log p_i^{(k)}$ \Comment{Entropy}
    \State $w_i\gets1-H_i/\log K$ \Comment{Reliability weight}
    \For{$c\in\mathcal{Y}$}
        \State $\tilde{p}\gets\mathrm{clip}(p_i^{(c)},\epsilon,1-\epsilon)$
        \State $\ell\gets\mathrm{clip}\!\left(\log\frac{\tilde{p}}{1-\tilde{p}},-M,M\right)$
        \State $\Score(c)\gets\Score(c)+w_i\ell$
    \EndFor
\EndFor
\State $\hat{Y}\gets\arg\max_{c\in\mathcal{Y}}\Score(c)$
\State \Return $\hat{Y}$  
\end{algorithmic}
\end{algorithm}

\noindent\textbf{Theoretical Analysis.}

\noindent\textbf{Proposition 1 (Information Preservation).}
\emph{Since the hard label $\hat y_i=\arg\max_k p_i^{(k)}$ is a deterministic function of
$\mathbf p_i$, the data processing inequality~\cite{cover2005elements} gives
$I(Y;\mathbf p_i)\geq I(Y;\hat y_i)$.
Thus, hard voting can discard reliability information even when two
segments support the same category with different confidence.}

TEF follows this principle by retaining the full distribution $\mathbf p_i$
instead of reducing each segment to $\hat y_i$.
Its entropy-based weight $w_i$ distinguishes decisive evidence from
ambiguous evidence, so two segments with the same hard label can contribute
different amounts.
The method therefore preserves the information discarded by majority voting
while producing an additive document-level score.

\noindent\textbf{Proposition 2 (Tempered Bayesian Pooling).}
\emph{For a binary contrast with equal priors, calibrated conditionally independent segment
posteriors satisfy the additive Bayes log-odds identity~\cite{varshney1997distributed}
$\log\frac{P(Y=c\mid D_{1:N})}{P(Y=\bar c\mid D_{1:N})}
=\sum_{i=1}^{N}\log\frac{P(Y=c\mid D_i)}
{P(Y=\bar c\mid D_i)}$.
Therefore, evidence is naturally accumulated on the log-odds scale rather
than through hard counts or raw probability averaging.}

TEF uses the one-vs-rest transform
$g(p)=\log\frac{p}{1-p}$ to implement this additive evidence scale.
Because LLM posteriors are approximate, it tempers each log-odds contribution~\cite{bissiri2016general}
with $w_i=(\log K-H(\mathbf p_i))/\log K$, the normalized information gain
from a uniform prior.
Near-uniform segments therefore have little influence, while low-entropy
stance-bearing segments retain a contribution close to the Bayesian
additive statistic.
This gives TEF its soft-censoring behavior, as uncertain sentences do not
accumulate by sheer number, while decisive sentences can determine the
final document-level measurement outcome even when they are few.

%% file: sections/4_benchmark.tex
\begin{figure*}[t]
\centering
\includegraphics[width=1\textwidth]{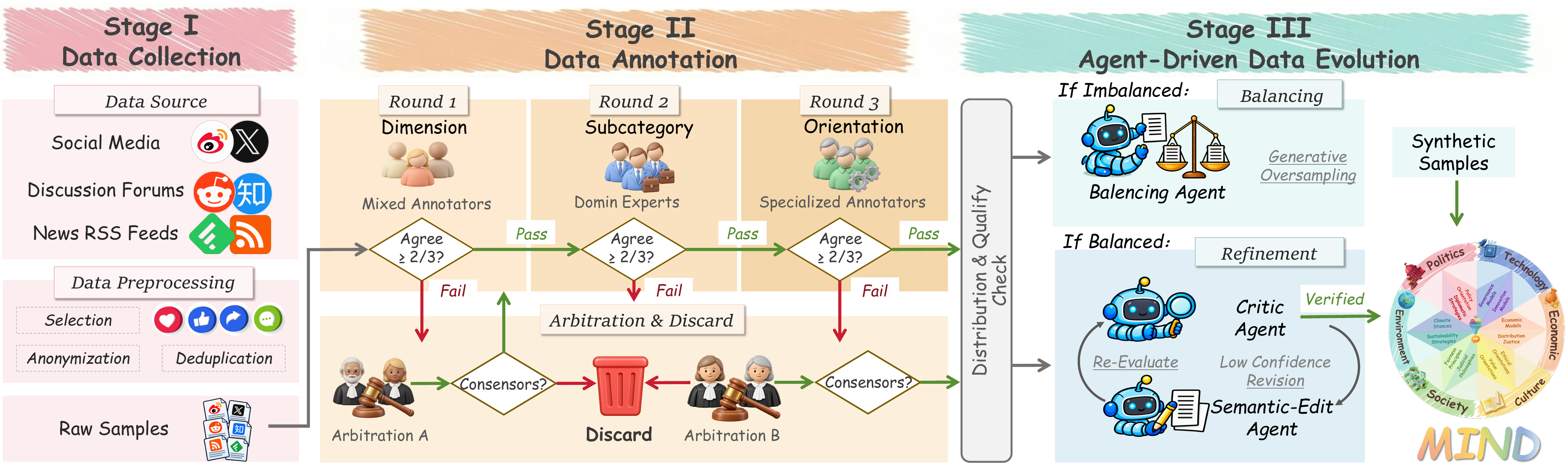}
\vspace{-0.6cm}
\caption{Construction pipeline of MIND. Posts are collected from public sources, filtered for relevance, 
anonymized, and deduplicated, then annotated in three rounds for dimension, subcategory, and orientation before being balanced and refined by LLM agents.}
\label{fig:pipeline}
\vspace{-0.6cm}
\end{figure*}

\begin{figure}[t]
\centering
\includegraphics[width=1.06\columnwidth]{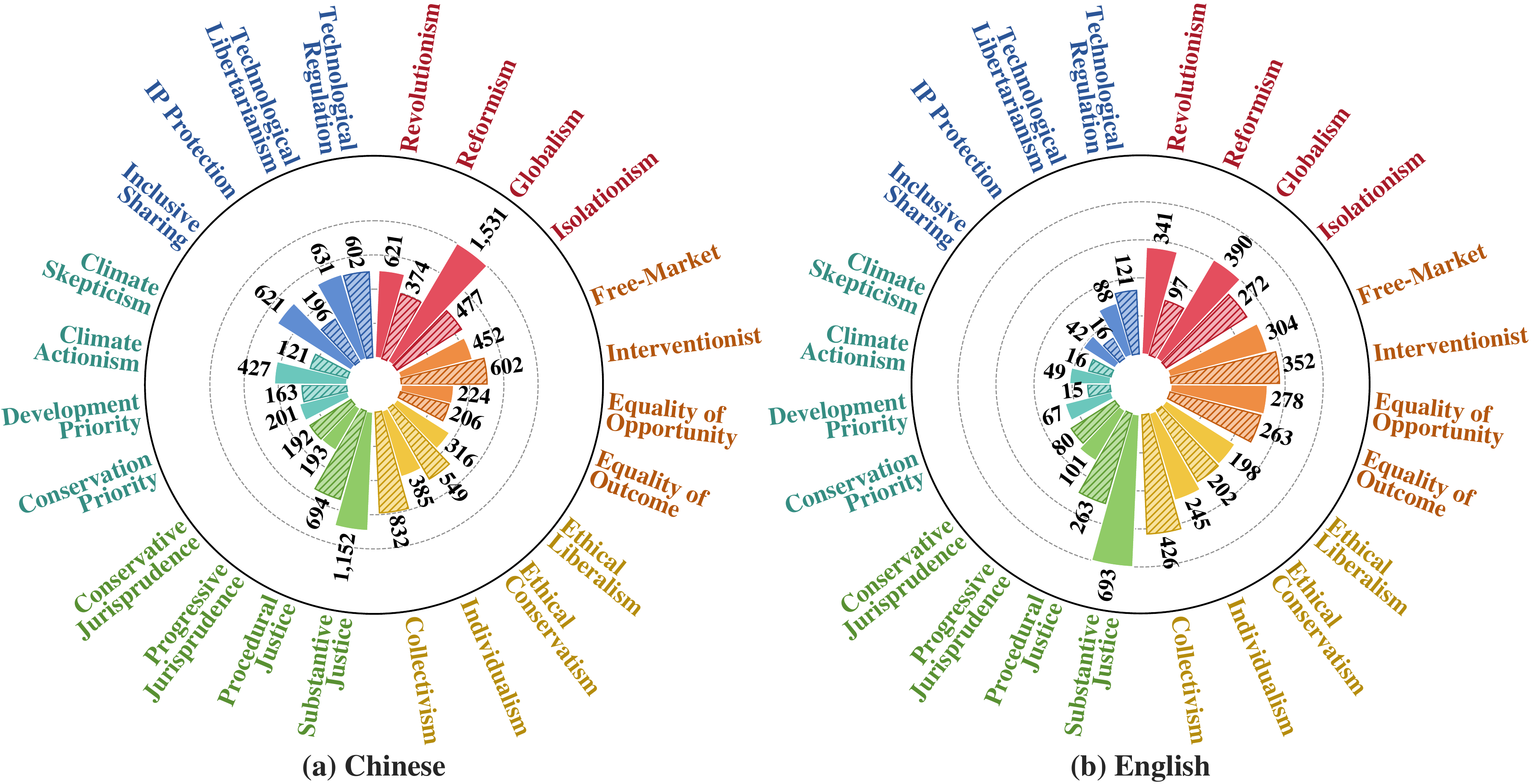}
\vspace{-0.6cm}
\caption{Statistics of MIND. Radial bars report the number of labels for each value 
orientation, with colors denoting dimensions and paired bars representing the opposing 
orientations of each subcategory. Panels (a) and (b) correspond to Chinese and English posts.}
\label{fig:statistics}
\vspace{-0.3cm}
\end{figure}
\section{The MIND Benchmark}
\label{sec:benchmark}

  
We construct the MIND dataset, a longitudinal, cross-lingual benchmark for measuring public value 
orientations. The final corpus contains 8,358 posts, including 5,474 Chinese posts and 
2,884 English posts over six value dimensions in Fig.~\ref{fig:statistics}. MIND is built through a three-stage pipeline. \noindent\textbf{Stage I: Data Collection.}
We collect candidate posts from X, Weibo, Reddit, Zhihu, discussion forums, and news feeds from January 2020 to January 2025, covering major 
public events such as the Russia--Ukraine war and the rise of generative AI. For each value dimension, keyword search is anchored to 
high-search-volume events. Raw posts are filtered for relevance and opinion content, anonymized, de-duplicated, and screened by 
an automatic quality score. \noindent\textbf{Stage II: Data Annotation.}
We adopt a three-round annotation framework to label the value dimension, subcategory, and orientation in sequence. Each round uses three annotators and accepts a label only when at least two agree. Failed or disputed cases enter arbitration, samples without a stable majority are discarded, and accepted samples pass distribution and quality checks before the evolution stage. \noindent\textbf{Stage III: Agent-Driven Data Evolution.}
To address class imbalance and semantic ambiguity, we introduce an agent-driven evolution stage. When a category is underrepresented, a balancing agent generates synthetic samples through generative oversampling. For low-confidence or ambiguous samples, a critic agent evaluates the alignment between text and label, and a semantic-edit agent revises the text to make the labeled orientation explicit. Only verified samples are retained in the final benchmark.

%% file: sections/5_experiments.tex
\vspace{-1.2cm}
\section{Experiments}
\label{sec:experiments}

\vspace{-0.3cm}
\noindent\textbf{Experiment Settings.} We evaluate three open-source models, Qwen2.5-7B~\cite{qwen25}, LLaMA3-8B~\cite{llama3}, and 
Qwen3-14B~\cite{qwen3}, and two proprietary models, DeepSeek-V3.2~\cite{deepseekv32} and GPT-4o-mini~\cite{gpt4omini}, 
using one shared prompt per language for TEF, Direct, Majority Vote, and Soft Vote.
Each prompt specifies the dimension and its two positions and requests a single answer token. We read the answer-token 
log-probabilities without sampling. Open-source models are served on RTX 3090 GPUs. We set $\epsilon=10^{-6}$ and $M=10$, 
and report dimension-level accuracy and macro-F1.

\noindent\textbf{Main Results.} In Table~\ref{tab:main}, 
TEF has the highest accuracy and macro-F1 for every model in both languages, exceeding the strongest baseline by 4.5 accuracy and 
4.6 macro-F1 points on average, so the gain comes from the fusion rule rather than from model-specific tuning. At the dimension level, 
TEF is the best or tied for best in 116 of 120 comparisons and never trails the strongest baseline by more than 2.6 points. 
Segmentation alone does not ensure improvement, since equal-weight voting, whether hard or soft, does not consistently improve on 
Direct. The gains of TEF concentrate where the baselines are weakest, on the smaller open-source models, on culture and politics 
in Chinese, and on environment in English, whereas on dimensions where Direct is already strong TEF matches it.


\noindent\textbf{Ablation Study.} Table~\ref{tab:ablation} evaluates TEF by removing one component at a time. On Qwen2.5-7B, both 
variants retain most of TEF's average gain over Soft Vote; on DeepSeek-V3.2, however, both fall below Soft Vote in both languages 
on average. Their complementarity reflects their distinct roles: the log-odds transform emphasizes decisive evidence, while 
entropy weighting suppresses uncertain sentences. The largest accuracy drop occurs on DeepSeek-V3.2's English environment 
dimension, where removing entropy weighting or the log-odds transform reduces accuracy by 22.9 or 23.8 percentage points.
\begin{figure}[htbp]  
\centering
\vspace{-0.3cm}
\includegraphics[width=0.96\columnwidth]{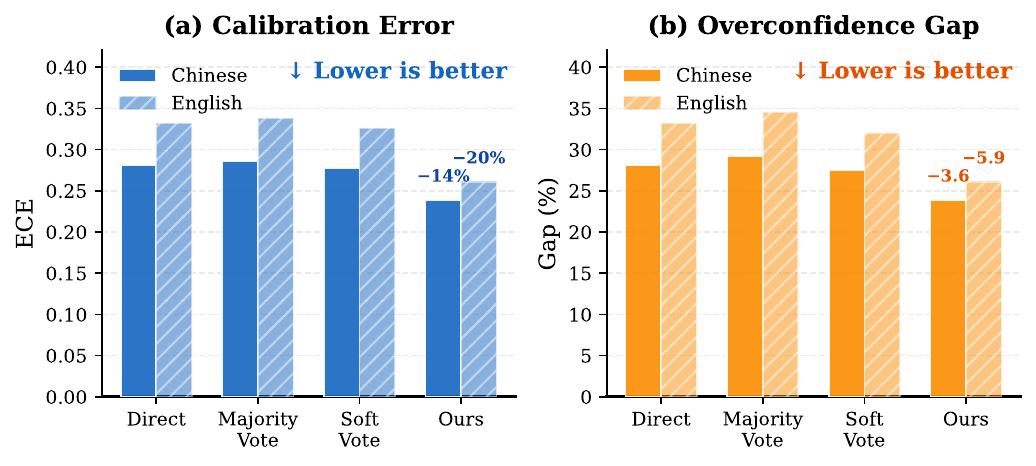}
\vspace{-0.5cm}
\caption{Calibration on Qwen2.5-7B. TEF achieves the lowest expected calibration error and 
overconfidence gap across both Chinese and English settings. Compared with Direct, Majority Vote, 
and Soft Vote, TEF aligns confidence with accuracy by tempering weak evidence.}
\label{fig:calib}
\vspace{-0.3cm}
\end{figure} 

\begin{table}[htbp]
\centering
\vspace{-0.425cm}
\caption{Robustness Analysis. Average accuracy (\%) over both languages under the original, verbose, and minimal prompts. Strongest baseline is the best of Direct, MV, and SV per prompt; Ours is TEF.}
\label{tab:prompt}
\vspace{0.35em}
\footnotesize
\setlength{\tabcolsep}{2pt}
\begin{tabular*}{\columnwidth}{@{\extracolsep{\fill}}l*{6}{c}}
\toprule
\multirow{2}{*}[\rulefix]{Model} & \multicolumn{3}{c}{Strongest baseline} & \multicolumn{3}{c}{Ours} \\
\cmidrule(lr){2-4}\cmidrule(lr){5-7}
& Original & Verbose & Minimal & Original & Verbose & Minimal \\
\midrule
Qwen2.5-7B & 61.5 & 60.8 & 59.2 & 70.2 & 69.8 & 68.6 \\
LLaMA3-8B & 60.8 & 60.0 & 58.2 & 67.0 & 66.4 & 65.3 \\
Qwen3-14B & 67.2 & 66.5 & 64.8 & 68.4 & 67.9 & 66.8 \\
DeepSeek-V3.2 & 65.0 & 64.3 & 62.7 & 69.1 & 68.5 & 67.5 \\
GPT-4o-mini & 67.2 & 66.5 & 65.0 & 69.3 & 68.8 & 67.8 \\
\bottomrule
\end{tabular*}
\vspace{-0.3cm}
\end{table}

\begin{table}[htbp]
\centering
\vspace{-0.42cm}
\caption{Efficiency Analysis. Total tokens and cost over the twelve evaluation subsets and mean time to first token (TTFT) on RTX 3090 GPUs. Voting covers MV and SV, which issue the same queries.}
\label{tab:efficiency}
\vspace{0.35em}
\footnotesize
\setlength{\tabcolsep}{4pt}
\begin{tabular*}{\columnwidth}{@{\extracolsep{\fill}}llccc}
\toprule
Model & Method & Tokens (M) & TTFT (ms) & Cost (\$) \\
\midrule
\multirow{3}{*}{Qwen2.5-7B} & Direct & 0.5 & 23.7 & 0.09 \\
 & Voting & 2.7 & 14.4 & 0.27 \\
 & \textbf{Ours} & 2.7 & 15.8 & 0.28 \\
\midrule
\multirow{3}{*}{LLaMA3-8B} & Direct & 0.6 & 28.6 & 0.10 \\
 & Voting & 3.1 & 15.2 & 0.28 \\
 & \textbf{Ours} & 3.1 & 17.3 & 0.30 \\
\bottomrule
\end{tabular*}
\end{table} 

\begin{strip}
\vspace{-0.25cm}
\centering
\refstepcounter{table}
\label{tab:main}
\parbox{\textwidth}{\textbf{Table \thetable}. Main results on MIND. Accuracy (Acc) and macro-F1 (F1) in percent for every model, language, and dimension. Cul., Eco., Soc., Pol., Tech., and Env. denote culture, economy, society, politics, technology, and environment. MV and SV denote Majority Vote and Soft Vote, and Ours denotes TEF. For each model, language, dimension, and metric, the best method is in \textbf{bold} and the second best is \underline{underlined}.}
\vspace{0.25em}

\scriptsize
\setlength{\tabcolsep}{1.2pt}
\renewcommand{\arraystretch}{0.9}
\begin{tabular*}{\textwidth}{@{\extracolsep{\fill}}ll*{12}{c}c*{12}{c}}
\toprule
& & \multicolumn{12}{c}{Chinese} & & \multicolumn{12}{c}{English} \\
\cmidrule(lr){3-14}\cmidrule(lr){16-27}
Model & Method & \multicolumn{2}{c}{Cul.} & \multicolumn{2}{c}{Eco.} & \multicolumn{2}{c}{Soc.} & \multicolumn{2}{c}{Pol.} & \multicolumn{2}{c}{Tech.} & \multicolumn{2}{c}{Env.} & & \multicolumn{2}{c}{Cul.} & \multicolumn{2}{c}{Eco.} & \multicolumn{2}{c}{Soc.} & \multicolumn{2}{c}{Pol.} & \multicolumn{2}{c}{Tech.} & \multicolumn{2}{c}{Env.} \\
\cmidrule(lr){3-4}\cmidrule(lr){5-6}\cmidrule(lr){7-8}\cmidrule(lr){9-10}\cmidrule(lr){11-12}\cmidrule(lr){13-14}\cmidrule(lr){16-17}\cmidrule(lr){18-19}\cmidrule(lr){20-21}\cmidrule(lr){22-23}\cmidrule(lr){24-25}\cmidrule(lr){26-27}
& & Acc & F1 & Acc & F1 & Acc & F1 & Acc & F1 & Acc & F1 & Acc & F1 & & Acc & F1 & Acc & F1 & Acc & F1 & Acc & F1 & Acc & F1 & Acc & F1 \\
\midrule
\multirow{4}{*}{Qwen2.5-7B} & Direct & \underline{44.6} & \underline{44.2} & \underline{44.6} & \underline{44.5} & \underline{74.4} & \underline{73.2} & \underline{55.2} & \underline{54.1} & 72.1 & \underline{66.3} & \underline{82.6} & \underline{81.8} & & \underline{54.0} & \underline{53.4} & 47.2 & \underline{47.1} & \underline{73.5} & \underline{69.7} & 56.6 & 53.6 & 62.9 & 56.2 & \underline{70.6} & \underline{64.6} \\
 & MV & 44.4 & 43.7 & 43.4 & 41.3 & 64.0 & 64.0 & 50.5 & 50.3 & 67.6 & 64.7 & 71.6 & 71.5 & & 53.1 & 52.2 & \underline{50.5} & 43.6 & 49.7 & 49.7 & \underline{60.9} & 53.3 & 62.9 & 58.1 & 51.4 & 48.9 \\
 & SV & 43.5 & 42.8 & 43.2 & 40.6 & 65.7 & 65.6 & 51.8 & 51.5 & \underline{69.9} & \underline{66.3} & 73.4 & 73.3 & & 53.2 & 52.5 & 49.0 & 41.8 & 50.8 & 50.8 & 60.8 & \underline{53.7} & \underline{63.4} & \underline{61.3} & 55.0 & 52.3 \\
 & \textbf{Ours} & \textbf{72.1} & \textbf{69.9} & \textbf{47.7} & \textbf{45.1} & \textbf{81.5} & \textbf{80.6} & \textbf{68.3} & \textbf{62.3} & \textbf{72.8} & \textbf{66.5} & \textbf{85.1} & \textbf{84.1} & & \textbf{62.5} & \textbf{58.2} & \textbf{59.4} & \textbf{59.4} & \textbf{79.6} & \textbf{77.3} & \textbf{67.8} & \textbf{66.4} & \textbf{64.6} & \textbf{61.6} & \textbf{80.7} & \textbf{67.2} \\
\midrule
\multirow{4}{*}{LLaMA3-8B} & Direct & \underline{63.5} & \underline{58.2} & 36.4 & 35.9 & \underline{73.8} & 70.7 & \underline{60.6} & \underline{56.5} & \textbf{70.9} & \underline{64.9} & \textbf{80.8} & \textbf{79.5} & & \underline{62.9} & \underline{53.4} & 41.1 & \underline{40.8} & \underline{74.8} & \underline{64.1} & 49.9 & 48.9 & \underline{54.9} & 52.6 & \underline{59.6} & \underline{56.6} \\
 & MV & 59.9 & 51.8 & 36.7 & 36.5 & 68.9 & 68.6 & 55.9 & 54.1 & 66.5 & 61.9 & 74.4 & 73.9 & & 54.9 & 48.9 & 47.8 & 35.3 & 60.2 & 59.7 & \underline{56.5} & \underline{50.4} & \underline{54.9} & \underline{54.4} & 48.6 & 47.6 \\
 & SV & 60.8 & 53.6 & \underline{37.3} & \underline{37.0} & 73.1 & \underline{72.9} & 54.3 & 52.8 & \underline{70.0} & 64.1 & \underline{78.2} & \underline{77.5} & & 55.4 & 49.0 & \underline{48.1} & 36.5 & 62.2 & 61.5 & 54.6 & 46.8 & 52.0 & 51.7 & 50.5 & 48.5 \\
 & \textbf{Ours} & \textbf{65.9} & \textbf{59.9} & \textbf{38.1} & \textbf{38.0} & \textbf{82.7} & \textbf{82.0} & \textbf{64.4} & \textbf{62.4} & \textbf{70.9} & \textbf{65.1} & \underline{78.2} & \underline{77.5} & & \textbf{63.3} & \textbf{56.2} & \textbf{55.8} & \textbf{53.5} & \textbf{80.5} & \textbf{77.6} & \textbf{67.5} & \textbf{63.4} & \textbf{58.9} & \textbf{58.7} & \textbf{78.0} & \textbf{70.5} \\
\midrule
\multirow{4}{*}{Qwen3-14B} & Direct & \underline{73.7} & 71.7 & \underline{41.3} & \underline{39.9} & \textbf{83.9} & \textbf{82.8} & 61.3 & 58.6 & \underline{74.5} & 68.9 & 87.9 & 86.8 & & 60.4 & 46.6 & \underline{56.2} & \underline{56.2} & 82.3 & 76.3 & \underline{50.1} & 46.3 & 60.0 & 53.6 & 74.3 & 60.1 \\
 & MV & 69.5 & 65.7 & \underline{41.3} & \underline{39.9} & 81.2 & 79.5 & 60.9 & 58.1 & 73.1 & 63.9 & 84.4 & 83.6 & & \underline{60.6} & 54.0 & \underline{56.2} & \underline{56.2} & \textbf{82.5} & \underline{76.7} & 49.7 & 46.1 & 57.1 & 52.6 & \underline{75.2} & \underline{63.4} \\
 & SV & \underline{73.7} & \underline{71.8} & 40.9 & 38.3 & \underline{83.7} & \underline{82.6} & 61.1 & 58.4 & \underline{74.5} & \underline{68.9} & \underline{87.9} & \underline{86.8} & & \underline{60.6} & 54.0 & 55.8 & 55.8 & 80.2 & 72.8 & 49.9 & \underline{46.5} & \textbf{60.6} & \underline{54.0} & \underline{75.2} & \underline{63.4} \\
 & \textbf{Ours} & \textbf{73.8} & \textbf{71.9} & \textbf{42.1} & \textbf{40.7} & \textbf{83.9} & \textbf{82.8} & \textbf{68.6} & \textbf{64.0} & \textbf{74.8} & \textbf{69.2} & \textbf{88.0} & \textbf{87.0} & & \textbf{62.3} & \textbf{54.3} & \textbf{57.9} & \textbf{57.8} & \textbf{82.5} & \textbf{76.8} & \textbf{50.7} & \textbf{47.4} & \textbf{60.6} & \textbf{54.1} & \textbf{76.1} & \textbf{64.2} \\
\midrule
\multirow{4}{*}{DeepSeek-V3.2} & Direct & \underline{68.7} & 66.0 & 39.6 & 39.4 & 79.8 & 78.8 & 62.3 & 59.8 & \underline{73.7} & 67.1 & \textbf{86.4} & 85.0 & & \underline{60.2} & \underline{59.9} & 49.5 & 49.5 & \underline{79.2} & \underline{76.9} & 55.7 & \underline{51.2} & 57.1 & 52.2 & 52.3 & 50.4 \\
 & MV & 68.5 & \underline{66.3} & 38.6 & 38.5 & 78.5 & 77.5 & 61.9 & 61.6 & 71.5 & 66.6 & 83.6 & 82.6 & & 59.0 & 57.9 & 57.7 & \underline{58.5} & 77.2 & 74.9 & \underline{61.8} & 49.2 & 57.1 & 54.5 & 54.1 & 52.7 \\
 & SV & 68.5 & 65.8 & \underline{40.0} & \underline{39.8} & \underline{80.2} & \underline{79.2} & \underline{62.6} & \underline{62.2} & 73.3 & \underline{67.6} & \underline{85.9} & \underline{84.9} & & 59.8 & 59.6 & \underline{58.1} & 58.0 & 79.1 & 76.8 & \textbf{62.7} & \underline{51.2} & \underline{58.3} & \underline{55.5} & 51.4 & 49.7 \\
 & \textbf{Ours} & \textbf{75.6} & \textbf{73.5} & \textbf{42.4} & \textbf{40.5} & \textbf{83.9} & \textbf{82.7} & \textbf{62.8} & \textbf{62.3} & \textbf{74.4} & \textbf{68.0} & \textbf{86.4} & \textbf{85.4} & & \textbf{63.7} & \textbf{60.3} & \textbf{58.8} & \textbf{58.8} & \textbf{83.9} & \textbf{80.3} & \textbf{62.7} & \textbf{55.4} & \textbf{60.0} & \textbf{58.7} & \textbf{74.3} & \textbf{66.4} \\
\midrule
\multirow{4}{*}{GPT-4o-mini} & Direct & 63.9 & 53.8 & \underline{42.2} & \underline{39.1} & \textbf{82.0} & \textbf{81.2} & 67.4 & 63.7 & \underline{74.9} & \underline{61.8} & \underline{83.6} & 82.3 & & \underline{60.9} & \underline{47.1} & \underline{49.8} & \underline{42.0} & \underline{81.7} & \underline{77.0} & 54.4 & 53.3 & \underline{63.4} & \underline{55.2} & 79.8 & \underline{68.7} \\
 & MV & 64.0 & \underline{54.4} & \underline{42.2} & 38.9 & 80.2 & 78.8 & \underline{67.7} & \textbf{65.3} & 73.0 & 60.4 & 82.0 & 80.1 & & 59.8 & 45.7 & \underline{49.8} & 40.3 & 79.7 & 75.6 & \underline{56.1} & \underline{55.5} & 61.7 & 53.9 & \underline{81.7} & 68.1 \\
 & SV & \underline{64.3} & 54.3 & 42.0 & 38.3 & 81.3 & \underline{79.9} & \underline{67.7} & 65.2 & 74.2 & 61.0 & \textbf{85.9} & \underline{84.3} & & 60.8 & 46.8 & \textbf{50.3} & \underline{42.0} & 81.3 & 76.8 & 55.8 & 54.5 & 62.3 & 54.9 & 80.7 & 68.5 \\
 & \textbf{Ours} & \textbf{74.0} & \textbf{71.2} & \textbf{42.9} & \textbf{41.1} & \underline{81.4} & \underline{79.9} & \textbf{67.8} & \textbf{65.3} & \textbf{75.3} & \textbf{68.9} & \textbf{85.9} & \textbf{84.4} & & \textbf{64.2} & \textbf{57.5} & \textbf{50.3} & \textbf{42.8} & \textbf{83.9} & \textbf{79.7} & \textbf{57.5} & \textbf{56.7} & \textbf{64.6} & \textbf{56.6} & \textbf{83.5} & \textbf{72.4} \\
\bottomrule
\end{tabular*}
\normalsize
\vspace{0.8em}

\refstepcounter{table}
\label{tab:ablation}
\parbox{\textwidth}{\textbf{Table \thetable}. Ablation of TEF. Accuracy and macro-F1 (\%) on Qwen2.5-7B and DeepSeek-V3.2; abbreviations as in Table~\ref{tab:main}. w/o Entropy keeps only the log-odds sum; w/o Log-Odds averages entropy-weighted probabilities instead of log-odds. Ours is the full TEF (\textbf{bold}).}
\vspace{0.05em}

\footnotesize
\setlength{\tabcolsep}{3pt}
\renewcommand{\arraystretch}{0.85}
\begin{tabular*}{\textwidth}{@{\extracolsep{\fill}}lll*{14}{c}}
\toprule
\multirow{2}{*}[\rulefix]{Model} & \multirow{2}{*}[\rulefix]{Language} & \multirow{2}{*}[\rulefix]{Method} & \multicolumn{2}{c}{Cul.} & \multicolumn{2}{c}{Eco.} & \multicolumn{2}{c}{Soc.} & \multicolumn{2}{c}{Pol.} & \multicolumn{2}{c}{Tech.} & \multicolumn{2}{c}{Env.} & \multicolumn{2}{c}{Average} \\
\cmidrule(lr){4-5}\cmidrule(lr){6-7}\cmidrule(lr){8-9}\cmidrule(lr){10-11}\cmidrule(lr){12-13}\cmidrule(lr){14-15}\cmidrule(lr){16-17}
& & & Acc & F1 & Acc & F1 & Acc & F1 & Acc & F1 & Acc & F1 & Acc & F1 & Acc & F1 \\
\midrule
\multirow{6}{*}[\rulefix]{Qwen2.5-7B} & \multirow{3}{*}{Chinese} & \textbf{Ours} & \textbf{72.1} & \textbf{69.9} & \textbf{47.7} & \textbf{45.1} & \textbf{81.5} & \textbf{80.6} & \textbf{68.3} & \textbf{62.3} & \textbf{72.8} & \textbf{66.5} & \textbf{85.1} & \textbf{84.1} & \textbf{71.3} & \textbf{68.1} \\
& & w/o Entropy & 71.3 & 68.8 & 45.6 & 42.9 & 79.9 & 79.1 & 67.4 & 61.4 & 71.7 & 65.8 & 81.6 & 80.7 & 69.6 & 66.5 \\
& & w/o Log-Odds & 71.8 & 69.6 & 46.2 & 43.5 & 80.3 & 79.5 & 67.2 & 61.1 & 72.2 & 66.2 & 81.5 & 80.6 & 69.9 & 66.8 \\
\cmidrule(lr){2-17}
 & \multirow{3}{*}{English} & \textbf{Ours} & \textbf{62.5} & \textbf{58.2} & \textbf{59.4} & \textbf{59.4} & \textbf{79.6} & \textbf{77.3} & \textbf{67.8} & \textbf{66.4} & \textbf{64.6} & \textbf{61.6} & \textbf{80.7} & \textbf{67.2} & \textbf{69.1} & \textbf{65.0} \\
& & w/o Entropy & 61.8 & 56.6 & 58.0 & 58.0 & 77.2 & 75.2 & 66.1 & 65.1 & 62.3 & 56.2 & 75.2 & 60.9 & 66.8 & 62.0 \\
& & w/o Log-Odds & 61.8 & 57.0 & 58.5 & 58.4 & 77.1 & 75.1 & 66.1 & 65.0 & 63.4 & 56.7 & 76.1 & 61.7 & 67.2 & 62.3 \\
\midrule
\multirow{6}{*}[\rulefix]{DeepSeek-V3.2} & \multirow{3}{*}{Chinese} & \textbf{Ours} & \textbf{75.6} & \textbf{73.5} & \textbf{42.4} & \textbf{40.5} & \textbf{83.9} & \textbf{82.7} & \textbf{62.8} & \textbf{62.3} & \textbf{74.4} & \textbf{68.0} & \textbf{86.4} & \textbf{85.4} & \textbf{70.9} & \textbf{68.7} \\
& & w/o Entropy & 68.1 & 65.4 & 39.2 & 39.0 & 79.6 & 78.7 & 62.4 & 61.9 & 72.7 & 67.1 & 85.7 & 84.8 & 68.0 & 66.2 \\
& & w/o Log-Odds & 68.5 & 65.7 & 39.3 & 39.1 & 79.5 & 78.5 & 62.7 & 62.2 & 73.3 & 67.6 & 84.9 & 83.8 & 68.0 & 66.2 \\
\cmidrule(lr){2-17}
 & \multirow{3}{*}{English} & \textbf{Ours} & \textbf{63.7} & \textbf{60.3} & \textbf{58.8} & \textbf{58.8} & \textbf{83.9} & \textbf{80.3} & \textbf{62.7} & \textbf{55.4} & \textbf{60.0} & \textbf{58.7} & \textbf{74.3} & \textbf{66.4} & \textbf{67.2} & \textbf{63.3} \\
& & w/o Entropy & 58.9 & 58.5 & 57.1 & 56.9 & 78.5 & 76.3 & 62.0 & 50.5 & 53.7 & 51.4 & 51.4 & 49.7 & 60.3 & 57.2 \\
& & w/o Log-Odds & 59.3 & 59.1 & 56.6 & 56.4 & 78.8 & 76.5 & 62.4 & 50.9 & 54.3 & 51.3 & 50.5 & 48.9 & 60.3 & 57.2 \\
\bottomrule
\end{tabular*}
\normalsize
\end{strip}


\noindent\textbf{Analysis.} 1) \emph{Calibration.} Because measurement outputs are thresholded or 
manually reviewed, document-level confidence should reflect correctness. In Fig.~\ref{fig:calib}, 
Direct and both voting rules are overconfident on Qwen2.5-7B, whereas TEF achieves the lowest 
expected calibration error~\cite{naeini2015obtaining,guo2017calibration} and overconfidence gap. Its confidence 
is therefore better suited to selective review. 2) \emph{Robustness.} To 
test whether the gain is due to the fusion rule rather than the shared prompt, we rerun all methods 
with a verbose prompt that adds role background and annotation guidelines and with a minimal prompt 
that retains only the classification directive. In Table~\ref{tab:prompt}, TEF remains best under 
every prompt and degrades the least under 
either perturbation. 3) \emph{Efficiency.} In Table~\ref{tab:efficiency}, TEF uses the same single-token 
queries as the voting methods, so its token count is identical. Its time to first token differs by only 
about two milliseconds. Relative to Direct, sentence-level methods cost roughly three times as much, 
but TEF is the only one that consistently turns this overhead into a performance gain.

%% file: sections/6_conclusion.tex
\vspace{-0.4cm}
\section{Conclusion}
\vspace{-0.2cm}
\label{sec:conclusion}

We cast long-text value measurement as the fusion of soft sentence 
decisions and derived TEF, which tempers each sentence's log-odds by 
its information gain. On MIND, TEF is more accurate 
than direct prediction and voting for five LLMs in two 
languages and better calibrated on Qwen2.5-7B, and its gain survives changes of the prompt.